\documentclass[conference]{IEEEtran}
\usepackage{times}

\usepackage[numbers]{natbib}
\usepackage{multicol}
\usepackage[bookmarks=true]{hyperref}
\usepackage{amsmath}
\usepackage{algorithm}
\usepackage{algpseudocode}
\usepackage{amssymb}
\usepackage{ upgreek }
\usepackage{graphicx}

\begin{document}

\title{cpRRTC: GPU-Parallel RRT-Connect for Constrained Motion Planning}

\author{Jiaming Hu*, Jiawei Wang*, Henrik Christensen}



%

\maketitle

\begin{abstract}
Motion planning is a fundamental problem in robotics that involves generating feasible trajectories for a robot to follow. Recent advances in parallel computing, particularly through CPU and GPU architectures, have significantly reduced planning times to the order of milliseconds. However, constrained motion planning—especially using sampling-based methods on GPUs—remains underexplored. Prior work such as pRRTC~\cite{huang2025prrtc} leverages a tracking compiler~\cite{thomason2024icra} with a CUDA backend to accelerate forward kinematics and collision checking. While effective in simple settings, their approach struggles with increased complexity in robot models or environments. In this paper, we propose a novel GPU-based framework utilizing NVRTC~\cite{nvrtc2019} for runtime compilation, enabling efficient handling of high-complexity scenarios and supporting constrained motion planning. Experimental results demonstrate that our method achieves superior performance compared to existing approaches. 
\end{abstract}

\IEEEpeerreviewmaketitle

\section{Introduction}
Many robotic tasks require constrained motion planning, where trajectories must not only avoid collisions but also satisfy task-specific constraints. These requirements significantly increase planning complexity, especially in high-dimensional spaces. Sampling-based approaches like CBiRRT~\cite{Dmitry2009cbirrt} address this by exploring the configuration space under constraints, but struggle in cluttered environments. To leverage GPU acceleration, cuRobo~\cite{curoboicra23} adopts an optimization-based approach for generating constrained motions efficiently. However, it lacks the global exploration guarantees provided by sampling-based methods. Recently, pRRTC~\cite{huang2025prrtc} introduces a GPU-accelerated RRT-Connect variant that achieves efficient exploration in high-dimensional, cluttered spaces, but it does not support constrained planning. This motivates the development of a GPU-based, sampling-driven framework that can handle both constraints and environmental complexity effectively.

During tree extension, pRRTC assigns one thread block to handle each motion, where each thread performs forward kinematics(FK) and collision checking(CC) for one waypoint of the motion. To reduce thread divergence, pRRTC uses a tracing compiler to generate CUDA code for both FK and CC. However, due to CUDA’s Single Instruction, Multiple Threads (SIMT) execution model, 
threads in a block must synchronize, meaning that even if one thread detects a collision early, it must wait for all other threads to complete their computations. In such cases, the motion is already invalidated, and the remaining FKs and CCs become redundant, leading to wasted computation. This inefficiency becomes especially pronounced when the number of robot collision spheres or environmental primitives is large. Consequently, pRRTC is typically limited to environments with fewer than 20 geometric obstacle primitives to maintain fast CC performance.

To address these limitations, we propose \textbf{cpRRTC}, an extension of pRRTC that supports constraint-satisfying motion generation and improves scalability in complex environments. Our key contributions are: (1) a parallel projection operator for constrained motion planning that fully utilizes GPU resources, and (2) integration of NVRTC-based runtime code generation with intra-block shared memory communication, allowing early termination upon collision detection. Together, these improvements enable cpRRTC to achieve efficient and scalable constrained motion planning in high-dimensional, obstacle-rich settings.

\section{Related Works}
RRT-Connect~\cite{kuffner2000rrtc} is an extension of RRT~\cite{lavalle2001rrt} designed to improve planning efficiency by simultaneously growing two trees. In each iteration, a random sample is used to extend one tree toward it, and then the other tree attempts to connect to the newly added node using a greedy expansion strategy. This bidirectional approach significantly increases the likelihood of rapidly finding a feasible path. Since individual tree extensions are independent, parallelization~\cite{curoboicra23, huang2025prrtc, thomason2024icra, le2024GTMP} offers a promising avenue for improving efficiency in sampling-based motion planning. VAMP~\cite{thomason2024icra} introduces vectorized motion validation using CPU-based Single Instruction, Multiple Data (SIMD), enabling data-level parallelism in collision checking. However, the scalability of CPU-based methods is inherently constrained by the limited number of available cores. To overcome this, pRRTC~\cite{huang2025prrtc} extends the core idea of VAMP to the GPU, leveraging its massively parallel architecture for greater computational throughput.

CBiRRT~\cite{Dmitry2009cbirrt} extends RRT-Connect by incorporating a projection operation that ensures all samples and interpolations remain on a constraint manifold, typically using iterative inverse kinematics solvers. However, the projection of a motion—i.e., a sequence of waypoints—must be performed sequentially, as each waypoint often depends on the result of the previous one. This sequential dependency limits opportunities for parallelism, as most threads must idle while waiting for the projection to complete. Consequently, directly applying parallelism to this process is inefficient.

\section{GPU-Parallel Project Operation on Motion}
As an extension of pRRTC for constrained motion planning, cpRRTC maintains a similar overall structure. Each tree extension is assigned to one thread block. Given a randomly sampled configuration $q_{\text{rand}}$ and its nearest neighbor $q_{\text{near}}$ in the tree, the thread block performs a steering operation to generate a new configuration $q_{\text{steer}}$ directed toward $q_{\text{rand}}$, subject to a predefined step size. An initial motion segment $\xi_{\text{init}}$, consisting of a fixed number of linearly interpolated states between $q_{\text{near}}$ and $q_{\text{steer}}$ (equal to the number of threads in the block), is computed and stored in shared memory. cpRRTC then projects $\xi_{\text{init}}$ onto the constraint manifold in parallel, keeping the start point fixed, to obtain a projected motion segment $\xi_{\text{projected}}$. If the projection succeeds, each thread performs forward kinematics (FK) and collision checking (CC) on its assigned intermediate state in $\xi_{\text{projected}}$. If the motion is entirely constraint-satisfying and collision-free, its final configuration $q_{\text{end}}$ is added to the tree with $q_{\text{near}}$ as its parent. The thread block then attempts to iteratively extend the tree from $q_{\text{end}}$ toward the nearest node $q_{\text{near-other}}$ in the opposite tree. In each iteration, an initial motion segment is generated, projected onto the constraint manifold, and evaluated. If the projected motion segment is collision-free and success to reduce the distance to $q_{\text{near-other}}$, its endpoint will be add to tree and continue to extend with the next motion segment. If the projected motion segment is not collision-free or fails to reduce the distance to $q_{\text{near-other}}$, the extension process terminates.

The following section describes two main differences from pRRTC. (1) Parallel projection on the given straight-line motion segment onto the constraint manifold; (2) how shared memory is utilized to enhance the efficiency of CC.

\subsection{Parallel Motion Projection}

\begin{figure}[t]
    \centering
    \includegraphics[width=0.45\columnwidth]{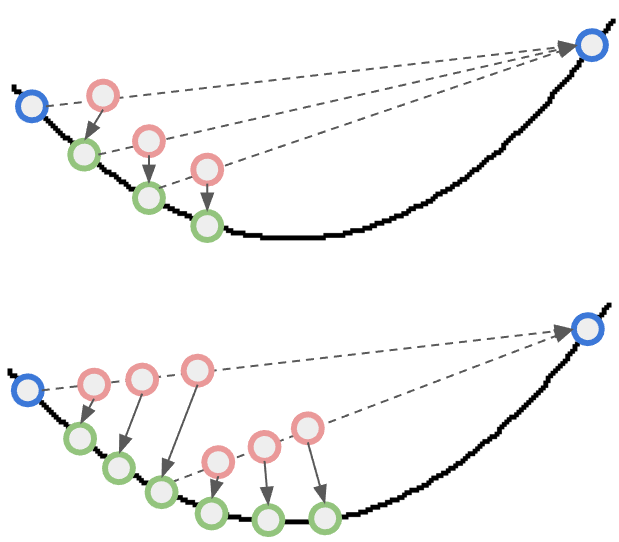}
    \caption{\textbf{Different ways to project motion segment} The curve line is the manifold, while the goal is projecting motion segment to manifold from one blue configurations to another. \textbf{Upper}: The original way to project motion segment in sequence. \textbf{Lower}: The parallel way to project motion segment.}
    \label{fig:parallelproject_vs_naiveproject}
\end{figure}

Compared to sequentially projecting each configuration onto the constraint manifold, the parallel motion projection operation projects all configurations of the motion segment simultaneously, allowing for more effective utilization of GPU resources. However, since each thread block can only handle a fixed number of waypoints, optimization methods like CHOMP~\cite{Nathan2009chomp}, which operate over trajectories of non-predefined size, are not directly applicable. Given an initial straight-line motion segment in the form
\[
\xi_{\text{init}} = [q_{0}, q_{1}, q_{2}, q_{3}, \ldots, q_{n}]^T,
\]
the parallel project function produces a constraint-satisfying motion segment:
\[
\xi_{\text{projected}} = [q_{0}, q'_{1}, q'_{2}, q'_{3}, \ldots, q'_{n}]^T.
\]
Here, $q_{0}$ remains fixed as the starting point, while the other configurations are iteratively projected in parallel. Note that the end configuration is not fixed. To ensure trajectory smoothness, the distance between consecutive configurations in the final motion must remain below a predefined threshold.

A naive approach would be to project all configurations in parallel independently. However, this can lead to discontinuities in the resulting projected motion segment. To address this, we incorporate an additional smoothing term into the project operation as shown in Alg.~\ref{alg:motoin_projection}.

\begin{algorithm} \label{alg:motoin_projection}
\begin{algorithmic}[1]
\Function{ParallelProject}{$\xi_{\text{init}}$, $tid$, $\tau_{task}$, $\tau_{sm}$}
    \State \text{shared memory } $\xi = \xi_{\text{init}}$
    \State \text{shared memory } $valid$
    \State \text{shared memory } $prog = 1$
    \State \text{shared memory } $isProj = False$
    \For{$i = 1$ to $\text{max\_iters}$}
        \If{$tid > prog \And tid <len(\xi)$}
            \State $E_{task} = \text{GetTaskError}(\xi[tid])$
            \State $J_{task} = \text{GetTaskJacobian}(\xi[tid])$
            \State $\nabla_{task} = J_{task}^{\dagger} E_{task}$
            \State $E_{sm} = abs(\xi[tid] - \xi[tid - 1]) - \tau$
            \State $J_{sm} = \xi[tid] - \xi[tid-1]$
            \State $\nabla_{sm} = J_{sm}E_{sm}$
            \State $\xi_{new}[tid] = \xi[tid] - \alpha(\nabla_{task} + \nabla_{sm})$
            \State $valid[tid] = E_{sm} < \tau_{sm} \And E_{task} < \tau_{task}$
        \EndIf
        \State syncthreads()
        \If{$tid = 0$}
            \For{$j=prog+1$ to $len(\xi)$}
                \If{valid[j]}
                    \State $prog = j$
                \EndIf
            \EndFor
            \If{$prog=len(\xi)-1$}
                \State $isProj = True$
            \Else
                \State $\xi = \xi_{new}$
            \EndIf
        \EndIf
        \State syncthreads()
        \If{isProj}
            \State return $\xi$
        \EndIf
    \EndFor
    \State \Return \textbf{None}
\EndFunction
\end{algorithmic}
\end{algorithm}

The \texttt{PARALLELPROJECT} function leverages thread-level parallelism to process the intermediate configurations of a motion segment $\xi_{init}$ in two main stages. In the first stage (lines 7--15), each thread computes the gradient for its assigned configuration, including both the task error $E_{task}$ and Jacobian $J_{task}$, as well as the smoothness error ${E_{sm}}$ and Jacobian $J_{sm}$ used to enforce trajectory continuity. At this stage, the intermediate waypoints for the updated motion $\xi_{new}$ are computed, but the original segment $\xi$ is not yet modified. At line 15, each thread checks whether its configuration $\xi[tid]$ satisfies the constraints and remains sufficiently close to its previous waypoint; the result is stored in the shared memory variable \texttt{valid}$[tid]$.

In the second stage (lines 18--29), a single thread evaluates whether the entire trajectory satisfies the constraints and maintains continuity by checking shared memory \texttt{valid}. A shared variable, $prog$, tracks how many initial configurations have been successfully projected; these configurations are left unchanged in subsequent iterations to improve stability. If the initial motion segment $\xi$ is not yet constraint-satisfying, $\xi_{new}$ is used to update $\xi$ at line 27. Finally, from lines 29--33, if the entire motion segment is constraint-satisfying and continuous, all threads exit the loop early.

\section{Intra-block communication of motion check}

Collision checking across threads is inherently parallel, as each thread evaluates an intermediate configuration independently. However, without communication among threads within a block, computation resources may be wasted---especially when an early collision renders the rest of the evaluation unnecessary. In pRRTC~\cite{huang2025prrtc}, the authors employed a tracking compiler to generate CUDA code for FK and CC. Due to limitations in the code generation process, it is difficult for FK and CC to leverage shared memory, preventing communication between threads during motion evaluation. As a result, all threads must complete both FK and CC, even when a collision is detected early by one thread. This inefficiency becomes especially problematic in cluttered environments with many obstacles, where exhaustive evaluation leads to significant performance degradation.

To address this, we propose using NVRTC to generate PTX code for the entire search algorithm rather than limiting it to FK and CC. This design enables the use of shared memory variables across FK and CC of threads within a block. When a thread detects a collision, it updates a shared memory flag, allowing other threads to terminate early. Because shared memory resides on-chip, it offers significantly lower latency compared to global or local memory, which are off-chip. Before performing primitive collision checks, each thread checks the shared flag; if a collision has already been reported, the current motion is marked invalid and further CC are skipped, improving overall efficiency.

\begin{figure}[t]
  \centering
  \includegraphics[width=0.65\columnwidth]{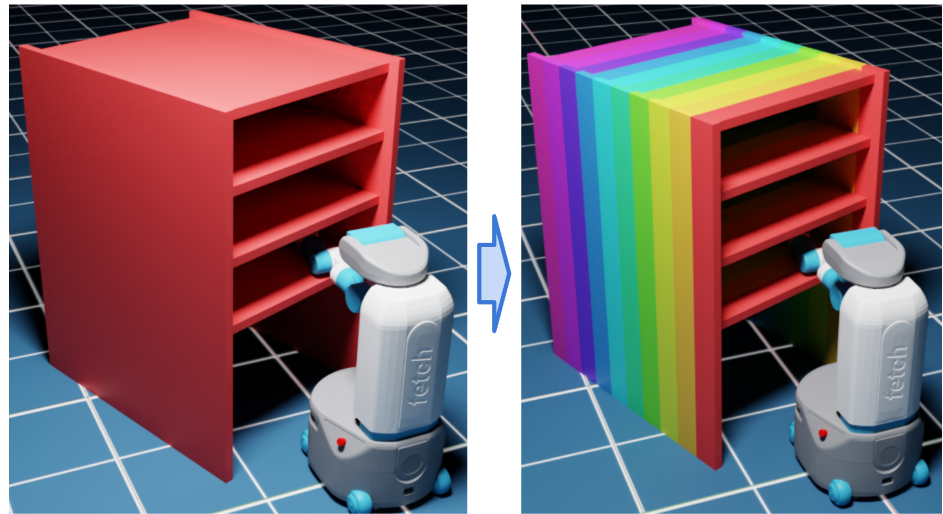}
  \caption{Illustrative example of the obstacle–segmentation procedure.  
           Left: original \emph{Bookshelf Tall} scene (1x).  
           Right: 10x version obtained by recursively subdividing each box. Each segmented box is represented by a different color.}
  \label{fig:scene_segmentation}
\end{figure}

\section{Evaluation}
This evaluation begin by evaluating the scalability of cpRRTC in environments with increasing obstacle density, followed by an assessment of its GPU-based constrained motion planning capabilities.

\subsection{Planning without Constraints}
We compare cpRRTC with pRRTC on MotionBenchMaker~\cite{thomason2024motions}. All planners were compiled using CUDA 12.8 and executed on a single workstation equipped with an NVIDIA RTX 5090 GPU. The planners sampled configurations from the Halton sequence~\cite{huang2025prrtc} and were reseeded only at the beginning of each trial. Our evaluation focuses on the Fetch and Franka robots, as they are commonly used platforms in prior work on cuRobo and pRRTC. To ensure a fair comparison, we follow the evaluation methodology used in pRRTC. To isolate the effect of obstacle count, we retain each scene’s overall geometry but subdivide every mesh into smaller axis-aligned boxes (Fig.~\ref{fig:scene_segmentation}).  
Three obstacle densities are tested: 1) the original scene, 2) every primitive split into 10x, and 3) every primitive split into 100x.

\begin{figure}[t]
    \centering
    \includegraphics[width=1.0\columnwidth]{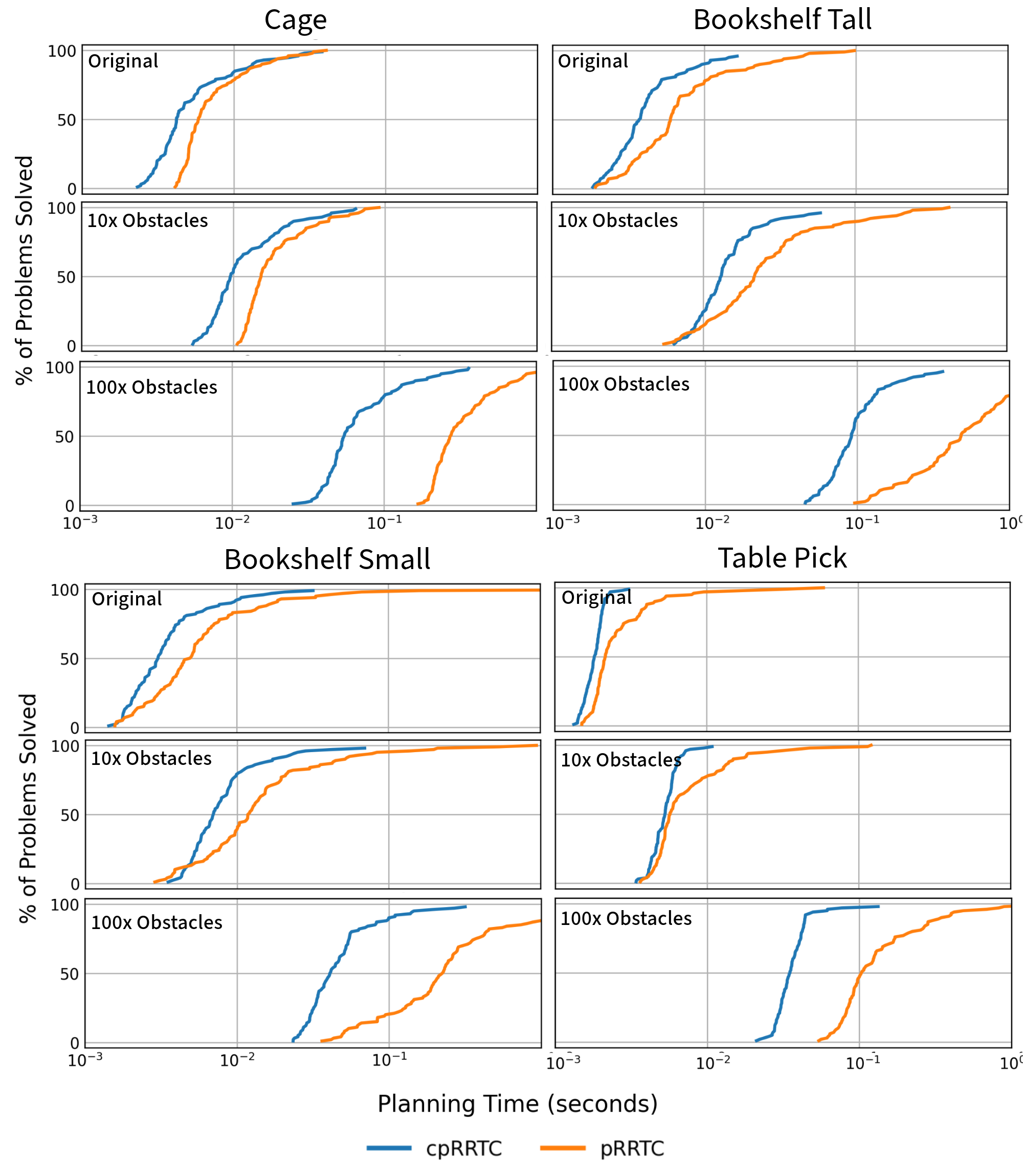}
    \caption{Cumulative distribution of solution times for the Fetch arm on MotionBenchMaker.}
    \label{fig:no_constraint_1_obs}
\end{figure}

Across the four tasks depicted in the original scenes (Fig.\ref{fig:no_constraint_1_obs}), the cpRRTC algorithm outperforms pRRTC with a speed-up of 2.8x. As the environmental complexity increases, this advantage expands significantly, achieving a speed-up of 3.4x at 10x complexity and approximately 6.6x at 100x complexity. These results demonstrate that our system scales significantly better than the prior approach.

\subsection{Planning with Constraints}
We evaluate cpRRTC-Parallel against cpRRTC-Naive, the algorithm with the sequential projection used in CBiRRT~\cite{Dmitry2009cbirrt}, and cuRobo \cite{curoboicra23}, the GPU-accelerated planner supporting constrained motion planning. Although cuRobo is optimization-based, it is therefore the de-facto reference for real-time constrained planning. Since cuRobo's default motion generation does not support joint-state configurations as planning goals, we utilize the \textit{TrajOptSolver} within the \textit{MotionGen} class to overcome this limitation.

\begin{figure}[t]
  \centering
  \includegraphics[width=1.0\columnwidth]{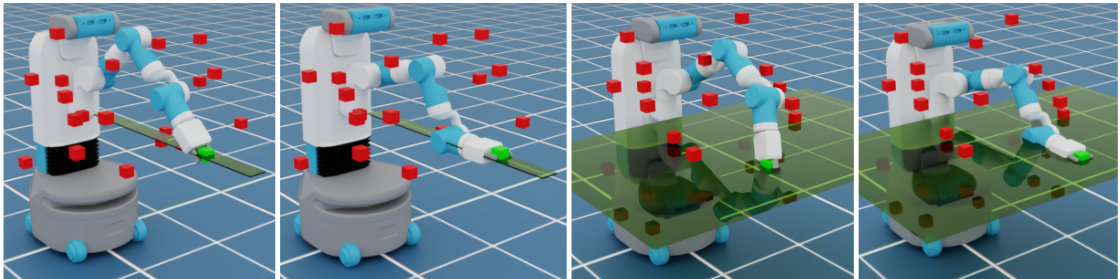}
   \vspace{-2em}
  \caption{Constrained motion planning task with obstacle(red cuboids). Green plane and line are constrained regions for end-effector.}
  \label{fig:constraint}
\end{figure}

CuRobo only enables constrained motion generation by allowing users to lock specific linear and angular axes, thereby solving motion along the remaining degrees of freedom. To fairly compare cpRRTC and cuRobo, we follow cuRobo’s constrained motion planning tutorial setup, adding randomly placed cuboid obstacles to increase complexity.
We consider two constraint variants: (1) a planar constraint where the end-effector is restricted to motion within a plane (i.e., one axis is constrained), and (2) a linear constraint where the end-effector is restricted to motion along a line (i.e., two axes are constrained). Representative examples are illustrated in Fig.~\ref{fig:constraint}. All evaluations use 100 randomly sampled start-goal joint-state pairs.

In general, across all benchmark settings, cpRRTC-Parallel solves the problems 165x times faster than cuRobo while improving success rate by 32\% on average. Compared with cpRRTC-Naive, the cpRRTC-Parallel is 1.1x faster and raises the success rate by 19\%.

\begin{figure}[t]
    \centering
    \includegraphics[width=1.0\columnwidth]{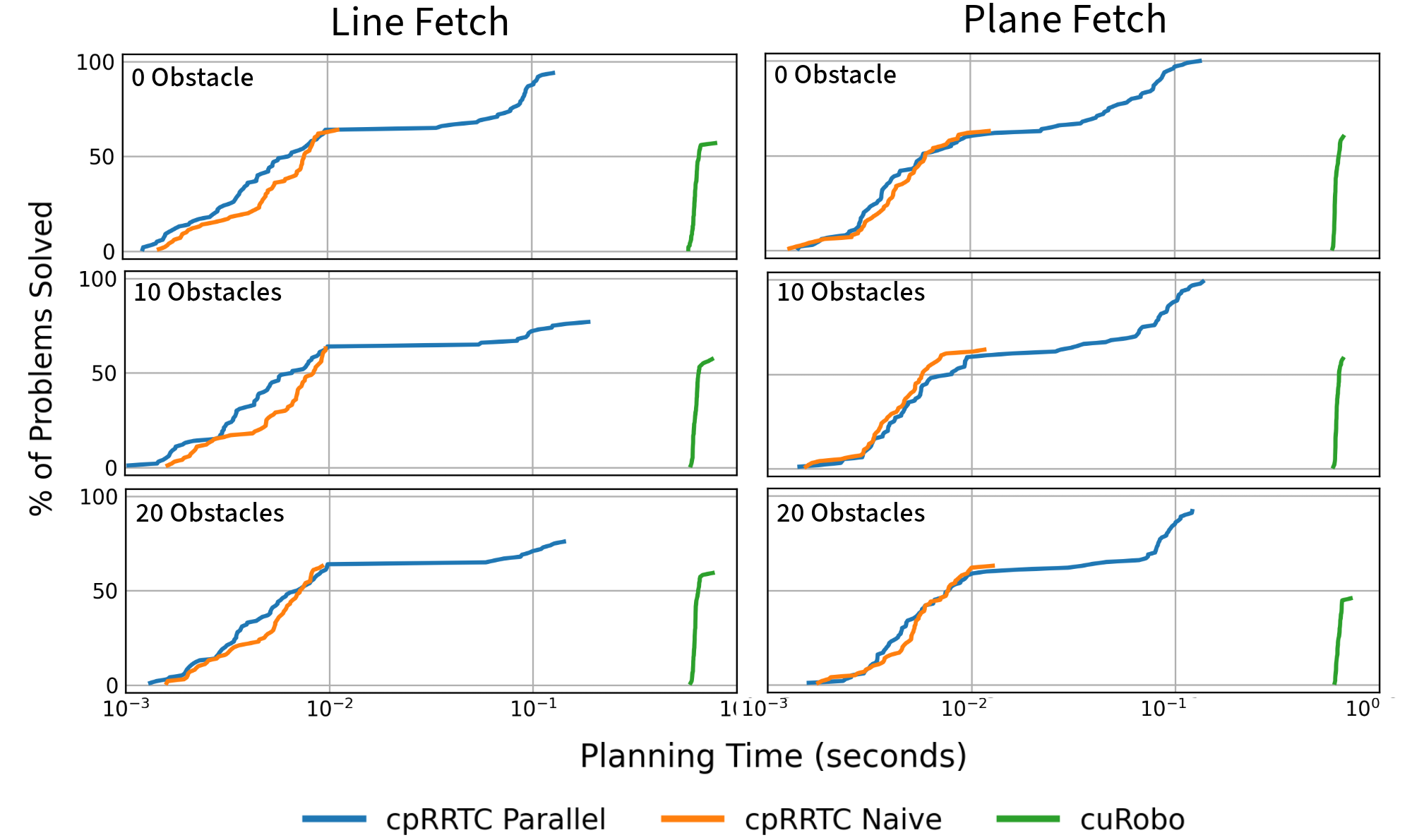}
    \caption{Cumulative distribution of solution times for the Fetch arm under line and plane constraint primitives.}
    \label{fig:constraint_0_fetch}
\end{figure}

For results with Fetch (Fig \ref{fig:constraint_0_fetch}), cpRRTC-Parallel achieved an average speedup of approximately 53× compared to cuRobo, while simultaneously increasing the mean success rate from 61\% to 90\%. Similar performance advantages were observed consistently across both constraint tasks. 
\begin{figure}[t]
    \centering
    \includegraphics[width=1.0\columnwidth]{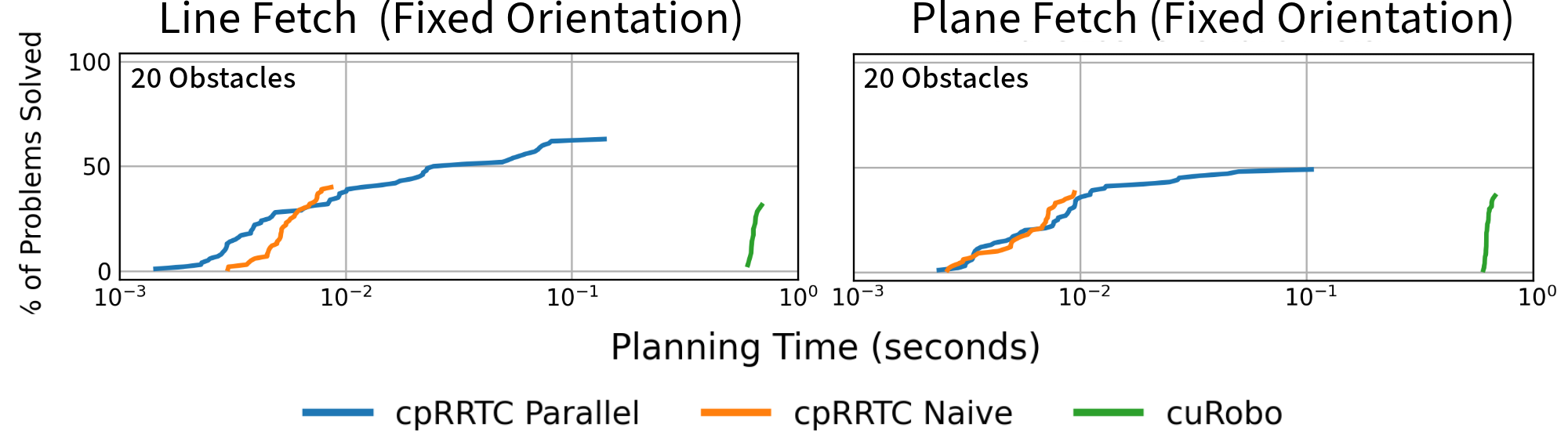}
    \caption{Cumulative distribution of solution times for the Fetch arm under line and plane with fixed orientation constraint primitives when the workspace has 20 obstacles.}
    \label{fig:constraint_lo_0_fetch}
\end{figure}
As the obstacle density increased, success rates for all planners dropped. Nonetheless, cpRRTC-Parallel consistently outperformed cuRobo under these challenging conditions, providing an average speedup of 52× and improving the success rate by approximately 22\%. For further evaluation, we impose fixed orientation constraints on the Fetch to increase task difficulty as shown in Fig. \ref{fig:constraint_lo_0_fetch}. In this case, cpRRTC-Parallel maintained comparable success rates to cuRobo while achieving approximately 120× faster planning times.

\begin{figure}[ht!]
    \centering
    \includegraphics[width=1.0\columnwidth]{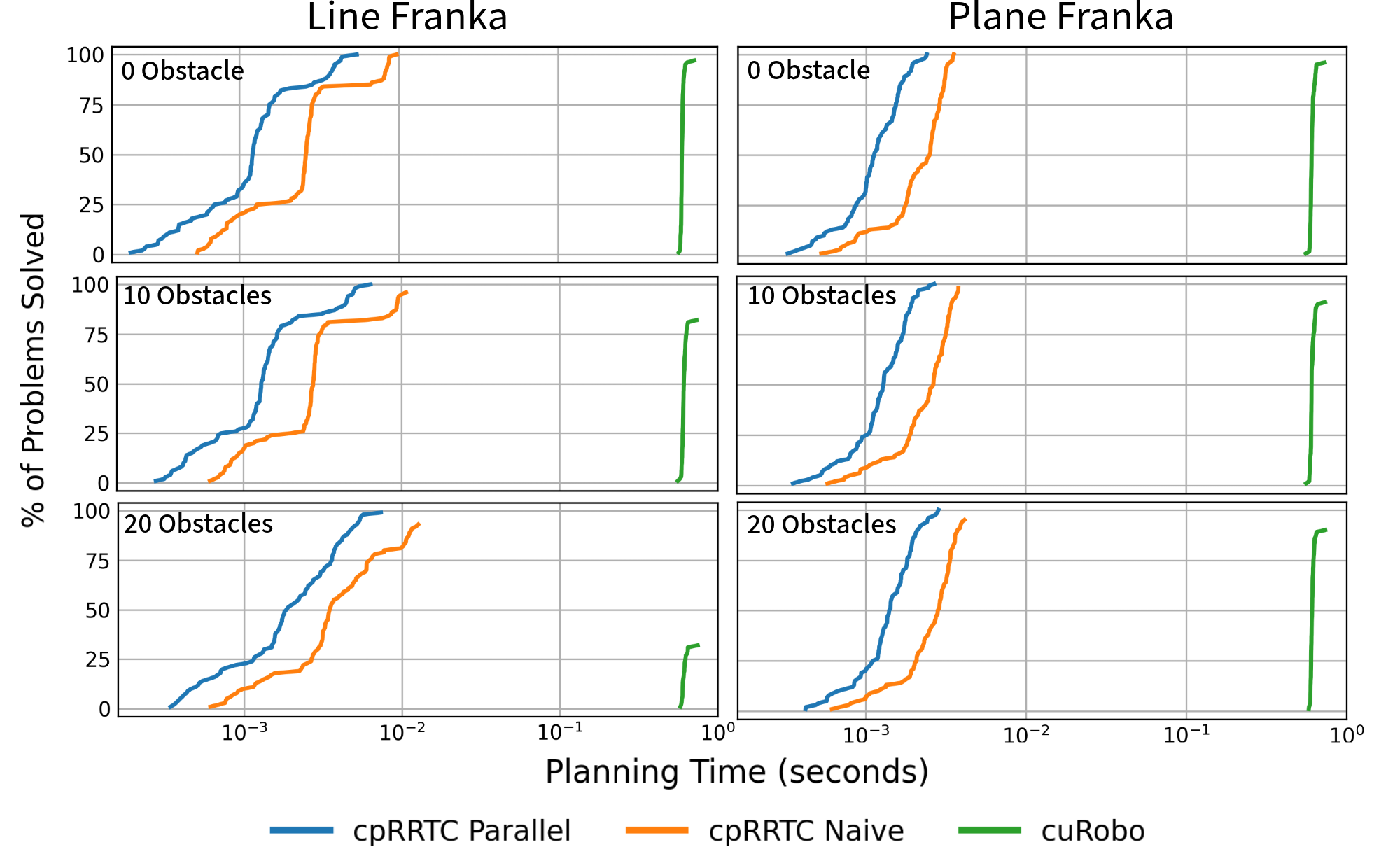}
    \caption{Cumulative distribution of solution times for the Franka arm under line and plane constraint primitives.}
    \label{fig:constraint_0_franka}
\end{figure}

For experiments with the Franka (Fig.~\ref{fig:constraint_0_franka}), cpRRTC demonstrated a similar performance advantage over cuRobo as observed with the Fetch. Notably, in the line-following task, as the number of obstacles increased, cpRRTC-Parallel maintained a consistently high success rate of approximately 99\%, whereas cuRobo’s success rate declined sharply to 32\%. Moreover, cpRRTC-Parallel achieved a 3.4x speed-up in planning time. We attribute this performance enhancement to the structure of the Franka (i.e it's slender arm and larger self-collision-free space) which simplified the planning problem and enable cpRRTC-Parallel to complete the planning task with fewer projection retries.




\newpage

\bibliographystyle{plainnat}
\bibliography{main}

\end{document}